\pdfoutput=1

\documentclass[11pt]{article}
\usepackage{authblk}
\usepackage[]{acl}

\usepackage{times}
\usepackage{latexsym}

\usepackage[T1]{fontenc}

\usepackage[utf8]{inputenc}

\usepackage{microtype}

\usepackage{adjustbox}
\usepackage{subfigure}
\usepackage{amsmath} 
\usepackage{multirow}
\usepackage{booktabs} 
\usepackage{xcolor,soul}
\usepackage{url}

\title{ProLex: A Benchmark for Language Proficiency-oriented \\ Lexical Substitution}

\author{\textbf{Xuanming Zhang}}
\author{\textbf{Zixun Chen}}
\author{\textbf{Zhou Yu}}
\affil{Computer Science Department, Columbia University}
\affil{\texttt {\{xz2995,zc2738,zy2461\}@columbia.edu}}

\begin{document}
\maketitle
\begin{abstract}

Lexical Substitution discovers $appropriate$ substitutes for a given target word in a context sentence. However, the task fails to consider substitutes that are of \textit{equal or higher proficiency} than the target, an aspect that could be beneficial for language learners looking to improve their writing. To bridge this gap, we propose a new task --- \textit{language proficiency-oriented lexical substitution}. We also introduce ProLex, a novel benchmark designed to assess systems' ability to generate not only appropriate substitutes but also substitutes that demonstrate better language proficiency. Besides the benchmark, we propose models that can automatically perform the new task. We show that our best model, a \textit{Llama2-13B} model fine-tuned with task-specific synthetic data, outperforms \textit{ChatGPT} by an average of 3.2\% in F-score and achieves comparable results with \textit{GPT-4} on ProLex.\footnote{Data and code available: \url{https://github.com/BillyZhang24kobe/LS_Proficiency}}

\end{abstract}


\begin{figure*}[h]
\centering
\includegraphics[width=1\textwidth]{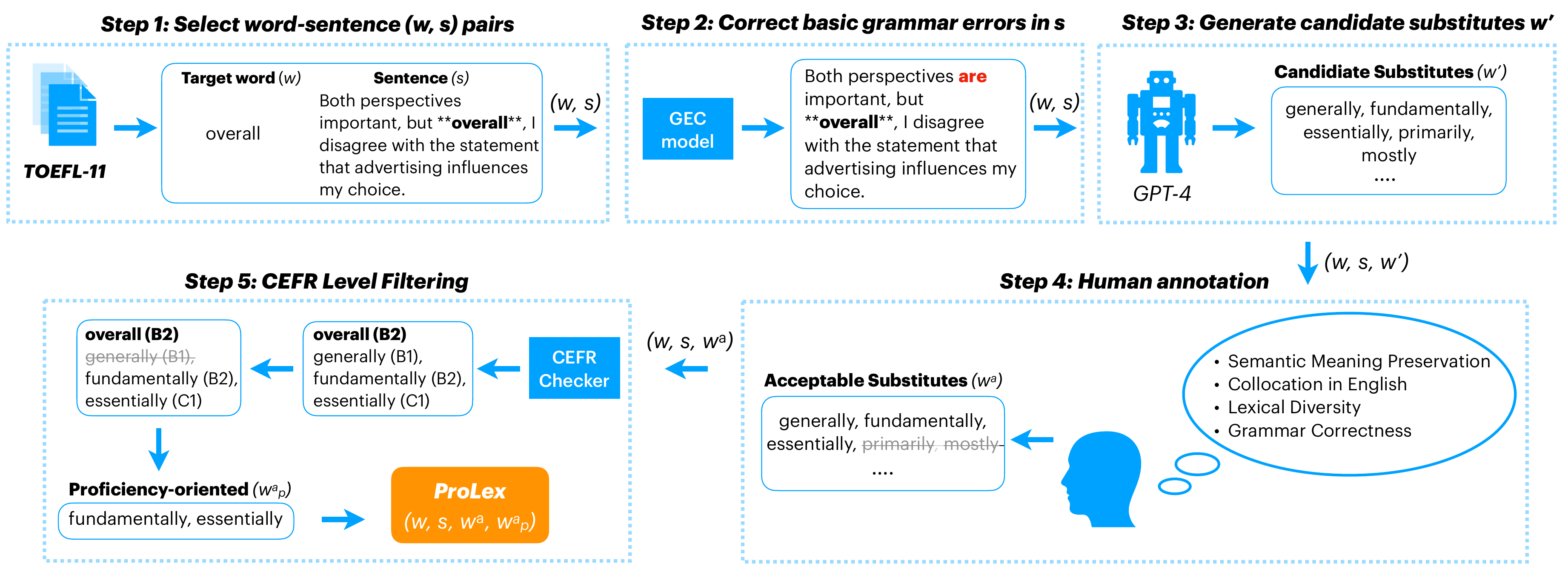}
\caption{The process of creating ProLex. We start by selecting word-sentence $(w, s)$ pairs from TOEFL-11 based on word frequency. Then we use a fine-tuned Grammar Error Correction (GEC) Model to correct basic grammar errors in the selected sentences. We use \textit{GPT-4} to generate candidate substitutes, each of which is denoted as $w'$. For each $(w, s, w')$ triple, we ask human expert to assess these $w'$ based on their appropriateness. The resulting list of accetpable substitutes is denoted as $w^a$. For all substitutes in $w^a$, we further apply a CEFR Checker \cite{cathovenai2023} to obtain their proficiency levels, and ultimately remove substitutes that demonstrate lower-level proficiency than the target word. This produces our final quadruplets in ProLex, namely $(w, s, w^a, w^a_p)$.} 
\label{fig:prolex_process}
\end{figure*}

\section{Introduction}

Nowadays, automatic English learning tools have become widespread across various educational settings. For instance, automatic grammar error correction systems \cite{omelianchuk2020gector,yasunaga2021lm,tarnavskyi2022ensembling,cao2023mitigating}, simplify grammar correction for learners and help enhance their writing skills. Besides grammar, enhancing vocabulary diversity is another crucial element in improving English writing \cite{smitherman2003language,johnson2016vocabulary,gonzalez2017contribution}. Nevertheless, English second-language (L2) learners frequently struggle to use a diverse vocabulary in their writing \cite{gu1996vocabulary,fan2020strategy,li2023exploring,sun2023affects}. They tend to use the same set of words repeatedly, which may negatively impact their performance on essay writing tests \cite{johnson2016vocabulary}.

The beginner L2 learners can leverage existing lexical substitution systems \cite{zhou2019bert,lacerra2021genesis,yang2022tracing, wada2022unsupervised,qiang2023parals} to enhance their vocabulary breadth. These systems are designed to identify contextually suitable substitutes for a target word, thereby assisting learners in discovering appropriate substitutions. However, merely knowing these contextually appropriate substitutes is insufficient for L2 learners. Prior benchmarks \cite{mccarthy2007semeval, kremer2014substitutes,horn2014learning,lee2021swords} for lexical substitution focus solely on the contextual appropriateness of generated substitutes. To enhance the vocabulary diversity and writing proficiency of English learners, guided by the principles of the zone of proximal development \cite{vygotsky1978mind}, we propose that the substitutions should also reflect an equal or higher level of language proficiency compared to the original target word. 


In this work, we present ProLex, a novel lexical substitution benchmark that evaluates system performances on \textit{language proficiency-oriented lexical substitution}, a new task that proposes substitutes that are not only contextually suitable but also demonstrate advanced-level proficiency. To construct ProLex, we begin by selecting target words and sentences according to their frequency in TOEFL-11 \cite{blanchard2013toefl11}, a comprehensive corpus of essays written by non-native English learners. This approach ensures a data distribution that is more representative of L2 English learners. Subsequently, to effectively harness human expertise in annotating advanced proficiency-oriented lexical substitutions, we adopt the methodology recommended by \citet{lee2021swords} --- directly having humans judge the appropriateness of candidate substitutes instead of generating them ad hoc. To facilitate this, we employ GPT-4 to generate an initial pool of substitute candidates. Then, the human annotators assess these candidates based on their contextual appropriateness using our language-learning-oriented annotation scheme. The proposed scheme takes into account the preservation of semantic meaning, common English collocations, lexical diversity, and grammatical correctness. We specifically recruited two annotators who are pursuing their PhDs in Linguistics, specializing in English language teaching and education. Following human annotation, we use the Common European Framework of Reference (CEFR) \cite{council2001common} to remove substitutions that have lower proficiency levels compared to the original target word.






In addition, we build models to perform \textit{language proficiency-oriented lexical substitution} automatically on ProLex. We also evaluate the performance of existing state-of-the-art (SOTA) lexical substitution systems, as well as prompting LLMs in zero-shot and in-context learning settings. The results of our experiments on ProLex reveal that our top-performing model, \textit{Llama2-13B}, which was instruction-tuned using a mix of synthetic and modified data, surpasses all existing SOTA lexical substitution systems by an average of 26.8\% in F-score. It also outperforms \textit{ChatGPT} by an average of 3.2\% and attains results on par with \textit{GPT-4}. We also demonstrate that, instruction tuning using task-specific synthetic data yields better results compared to zero-shot LLMs on ProLex.



Overall, our contributions are threefold:
\begin{itemize}
    \item We propose a new task, namely \textit{language proficiency-oriented lexical substitution}, to help L2 English learners enhance their vocabulary diversity and writing proficiency.
    \item We present a novel benchmark ProLex to assess systems' ability to perform the task by generating substitutes that are not only contextually appropriate but also reflect advanced language proficiency levels.
    \item We fine-tuned models with task-specific synthetic data and evaluated them with ProLex. The models attain results that are comparable to, or better than, the out-of-the-box LLMs, such as \textit{GPT-4} and \textit{ChatGPT}.
\end{itemize}

\section{Related Work}
We first outline the task of lexical substitution, then introduce recent high-quality benchmark in this field, Swords, as detailed in \cite{lee2021swords}. Next, we describe TOEFL-11 \cite{blanchard2013toefl11}, a large-scale corpus of written essays composed by non-native English learners, which we adopt as the corpus for selecting context sentences and target words for ProLex.


\paragraph{Lexical Substitution} Lexical substitution is a well-established task, which was originally defined in \citet{mccarthy2002lexical} - given a target word $w$ in a given context $c$, one needs to generate a list of substitutes $w'$ that can be used to replace $w$ in $c$. Specifically, the context $c$ refers to one or more sentences encompassing the target word $w$. The target word itself, which may be selected manually by human annotators \cite{mccarthy2007semeval}, or automatically based on its part-of-speech as discussed in \citet{kremer2014substitutes,lee2021swords}, is a single word within this context. The substitute $w'$ for the target word can be either a single word or a phrase. Although $w'$ can be ungrammatical in previous benchmarks, we want to incorporate only grammatically correct substitutes in ProLex. Moreover, the past benchmarks do not consider the language proficiency of the substitutes $w'$. In ProLex, our approach includes substitutes that are not only acceptable but also demonstrate equal or higher proficiency level compared to the target word.



\paragraph{Swords} Different from other popular benchmarks \cite{mccarthy2007semeval,kremer2014substitutes} depending on recall from humans as the only source of data, Swords \cite{lee2021swords} enhanced the collection of data in lexical substitution by treating it as a classification problem. This method was guided by the intuition that it is easier for humans to judge the appropriateness of given substitutes than to create them spontaneously. This approach resulted in a higher-coverage and higher-quality benchmark. In our human annotation process, we follow Swords's established instructions, and propose a language-learning-oriented annotation scheme for the appropriateness judgement.


\paragraph{TOEFL-11 Essay Corpus}  The large-scale non-native English writing corpus, TOEFL-11 \cite{blanchard2013toefl11}, contains writing samples from non-native TOEFL test takers from 2006 to 2007. In particular, it contains $12,100$ essays, each of which is categorized into one of the three score levels: high, medium, and low. Focusing on enhancing the writing skills of English learning beginners, our methodology involves selecting context sentences and target words primarily from essays with \textit{medium} and \textit{low} scores. This approach is based on the assumption that beginner L2 English learners are likely to possess a more limited vocabulary range \cite{fan2000big}.



\section{ProLex Benchmark}
We propose ProLex, a benchmark for language proficiency-oriented lexical substitution. ProLex is composed of quadruplets, each containing a target word, a context sentence, a list of acceptable substitutes, and a list of proficiency-oriented substitutes ($w$, $s$, $w^a$, $w^a_p$). As indicated by \citet{lee2021swords}, previous benchmarks \cite{mccarthy2007semeval, kremer2014substitutes} gathered substitute words by prompting humans to generate them ad hoc. However, many viable substitutes, challenging for humans to devise, might ultimately be overlooked. Therefore, we follow \citet{lee2021swords} by first obtaining a set of candidate substitutes and later asking annotators to judge their appropriateness. Formally, given a context $s$, target word $w$, and candidate substitute $w'$, the annotators are asked to judge whether $w'$ is an acceptable substitute for the target word: ($w$, $s$, $w'$) $\to$ \{1, -1, 0\},  with 1, -1 and 0 indicating "accept", "reject", and "uncertain" respectively. Figure \ref{fig:prolex_process} shows the complete benchmark creation process for ProLex. In the following sections, we will first illustrate the data creation process in Section \ref{sec:data_col}. Then, in Section \ref{sec:ann_main}, we will describe our language-learning-oriented annotation scheme. In Section \ref{sec:cefr_filter}, we show how we filter substitutes based on CEFR proficiency levels.




\subsection{Data Creation}
\label{sec:data_col}
Since we focus on improving the writing proficiency of beginner L2 learners, we turn to TOEFL-11 \cite{blanchard2013toefl11}. We will select contexts, target words, correct basic grammar errors, and generate candidate substitutes through the following three steps.



\paragraph{Step 1: Select contexts, and target words.}

The TOEFL-11 dataset comprises $12,100$ essays, each scored as high, medium, or low. Specifically, there are $4,202$ high-scored essays, $6,568$ with medium scores, and $1,330$ categorized as low. Given the limited vocabulary range of low and medium-proficiency L2 English learners, we intend to select the words that are frequently used in their essays as the target words in our benchmark. For each essay of interest, we select mainly the content words (i.e. nouns, verbs, adjectives, and adverbs) that are used \textit{at least three times} in the essay. Then, for each such target word, we randomly choose one sentence from all sentences containing that target word. The general statistics of all target words and context sentences are presented in Appendix \ref{sec:stat_tword_w} Table \ref{tab:stat_tword_sel}. In sum, we selected $2,531$ unique word-context $(w,s)$ pairs from low and medium-level essays from TOEFL-11 corpus.

\paragraph{Step 2: Correct basic grammar errors in selected context sentences.}
We observed that the chosen sentences frequently contain fundamental grammar mistakes, such as spelling errors, which can obscure the intended meaning of the context. Example grammar errors are demonstrated in appendix \ref{app:grammar_error}. To correct these errors, we follow \citet{rothe2021simple} and fine-tune a GPT-2 model \cite{radford2019language} on CLang-8 dataset.\footnote{https://github.com/
google-research-datasets/clang8} The experiment details are also shown in appendix \ref{app:grammar_error}. The resulting model is deployed to correct the basic grammar errors in all of our $2,531$ selected context sentences $s$. We then sample $1,125$ $(w, s)$ pairs for human annotations.\footnote{We used $125$ pairs for calculating inter-annotator agreement and $1,000$ for the annotation. We will include more pairs to expand the benchmark in future work.}


\paragraph{Step 3: Generate candidate substitutes.}
Powered by recent advances in LLMs \cite{brown2020language,wei2022chain}, for each ($w$, $s$) we generate the candidates using GPT-4. Specifically, we prompt GPT-4 with five in-context examples to generate exactly five candidate substitutions for each of our $1,000$ ($w$, $s$) pairs --- thus resulting in $5,000$ $(w, s, w')$ triples, with $w'$ representing each candidate substitute for each $(w, s)$ pair. Figure \ref{fig:prolex_process} demonstrates an example of the generated candidates. The complete prompt for GPT-4 generation is shown in Appendix \ref{sec:gpt4_prompts}.




\subsection{ProLex Annotation Scheme}
\label{sec:ann_main}
Different from previous schemes \cite{mccarthy2007semeval, kremer2014substitutes, lee2021swords}, ProLex deems a substitute valid if it maintains the original sentence's meaning, aligns with common English usage, and is grammatically correct. Furthermore, ProLex encourages a diverse set of substitutes as it can expand the vocabulary of language learners. A detailed description of the annotation scheme is in Appendix \ref{sec:ann_scheme}. Formally, as described in the previous section, for each $(w, s, w')$ pair, we ask the annotators to judge whether $w'$ is an acceptable substitute for $w$, by assigning one of the following classes to $w'$, 1 (acceptable), -1 (unacceptable) and 0 (uncertain). Subsequently, for each $(w, s)$ pair, we will combine all $w'$ that are labeled as acceptable to form $w^a$, leading to $(w, s, w^a)$ triples. In the following, we will discuss the features of the annotation scheme in detail.




\subsubsection{Semantic Meaning Preservation}
Given a triple $(w, s, w')$, namely a target word $w$, and the context sentence $s$ where $w$ is situated, the word substitute $w'$ should preserve the semantic meaning of the original context sentence $s$. Concretely, there are three cases of acceptable substitutes (i.e. labeled as "1") for meaning preservation in ProLex. Firstly, the annotators accept a substitute of $w$ if they convey exactly the same meaning in $s$. Besides, a substitute is also acceptable if it is an entailment from the sense of the target word. For example, \textit{trust a person} entails \textit{rely on a person} in the following sentence: \textit{"I **trust** a person who has more knowledge than I do."} Hence, \textit{rely on} is a valid substitute to target word \textit{trust} in the example. Lastly, a substitute can be used figuratively and convey a similar meaning as the target word $w$. In the following sentence: \textit{"Once the undergraduate studies are **pursued** by a student, the student is more aware of different subjects and the knowledge he has gained in the period of his studies."}, \textit{embarked on} is an acceptable substitute as it is a figurative use case, preserving the meaning of the target word \textit{pursued}. 


\subsubsection{Common Collocations in English}
Learners can improve their general levels of writing proficiency in the language through collocation knowledge \cite{howarth1998phraseology,gitsaki1999second,boers2018teaching}. In ProLex, annotators accept substitutes only if they are common English collocations within the context sentence. For instance, consider the following example where the target word is \textit{defying}: \textit{"Successful people are not only seeking new experiences, **defying** obstacles and hardship but also trying to be creative."} Suppose we have a candidate substitute \textit{conquering}, even though \textit{conquering obstacles} conveys a similar meaning as \textit{defying obstacles}, it will be rejected since it is not a common expression and collocation in English. 

In addition, in cases where the annotators are unsure about the collocation of any expression, we encourage them to refer to an external collocation knowledge base, such as COCA, the Corpus of Contemporary American English \cite{davies2010corpus}, which contains 400 million words from various genres (e.g. spoken, fiction, newspapers and etc.) distributed from 1990 to present time. In particular, we ask the annotator to accept certain expressions if the frequency\footnote{https://www.english-corpora.org//coca/} of it queried from COCA is greater than $5$. Take \textit{conquer obstacles} from the previous example, since there is only one instance of \textit{conquer obstacles} in COCA, it is rejected from the candidate list.

\subsubsection{Lexical Diversity}
A richer and more diverse vocabulary can improve both the quality and effectiveness of communication in English \cite{yu2010lexical}. To enhance lexical diversity in ProLex, annotators are asked to mark a substitute as acceptable if, in general, it matches at least one connotation of the target word in the context sentence. For instance, in the sentence \textit{"In fact they not only give the possibility to move in a short time but sometimes they are also **precious** goods."}, \textit{precious} conveys a connotation of high value in price or importance. Hence, the following candidates are all acceptable: \textit{valuable, prized, cherished, invaluable} and \textit{treasured}, as they express either "high value in price" or "importance", matching at least one connotation of the target word \textit{precious}.


\subsubsection{Grammar Correctness}
Although previous benchmarks \cite{mccarthy2007semeval, kremer2014substitutes, lee2021swords} do not involve grammar correction, ProLex considers the substitutes to be acceptable only if they are grammatically correct. For instance, in the sentence \textit{"Nevertheless, who are mostly responsible for these **research**?"}, all of the following candidates are rejected because none of them are in plural forms: \textit{study, investigation, analysis, examination} and \textit{inquiry}.



\subsection{Substitutes Filtering based on CEFR Proficiency Levels}
\label{sec:cefr_filter}
Taking the triples $(w, s, w^a)$ from human annotations in section \ref{sec:ann_main}, for each target word $w$ in a context sentence $s$, and the list of acceptable substitutes $w^a$, we perform filtering to select substitutes with equal or higher proficiency compared to the target word to form $w^a_p$ --- resulting in the final quadruplets of ProLex $(w, s, w^a, w^a_p)$. Concretely, we refer to the CEFR Checker \cite{cathovenai2023} to automatically determine the CEFR level of each target word $w$ and its associated acceptable substitutes $w^a$. An example filtering process is shown in Figure \ref{fig:prolex_process}. Note that \textit{generally} is removed from the set since its CEFR level (i.e. B1) is lower than that of \textit{overall} (i.e. B2).

\begin{table}
\centering
\begin{adjustbox}{max width=\linewidth}
    \centering
    \begin{tabular}{ccccc} \toprule 
         \textbf{Essay Proficiency}&  \textbf{\# ($w$, $s$)}& \textbf{Avg $s$ length} & \textbf{\# $w^a$} & \textbf{\# $w^a_p$}\\ \midrule 
         low&  169&  18.8& 489& 419 \\ 
         medium&  511&  22.7& 1466 & 1140\\ 
         all&  680&  21.8& 1955 & 1559\\ \bottomrule
    \end{tabular}
\end{adjustbox}
\caption{Statistics of ProLex, separated according to essay proficiency. "all" denotes the combinations of both "low" and "medium"-level essays.}
\label{tab:stat_test}
\end{table}

\section{Dataset annotation process and statistics}
Considering the necessity for substantial semantic understanding and knowledge of English collocations in the annotation process, we recruited two annotators currently pursuing PhDs in Linguistics, with specializations in English language teaching and education. On the $125$ $(w, s)$ pairs, namely $625$ $(w, s, w')$ triples, the two annotators reached an inter-annotator agreement of $\kappa = 0.60$, indicating a near-substantial level of agreement.\footnote{We perform Cohen's Kappa \cite{cohen1960coefficient} to calculate the inter-annotator agreement. We exclude the labels with 0 to consider only the contexts that are conceivable to both annotators (i.e. labeled as 1 or -1).} Then, they annotated the complete test set separately, with each annotating $2,500$ $(w, s, w')$ triples sampled from section \ref{sec:data_col}. 
In total, among all $5,000$ candidate substitutes $w'$, $39\%$ is labeled as 1 (accept), while $51\%$ and $10\%$ are annotated as -1 (reject) and 0 (uncertain), respectively. This entails that GPT-4 generated candidates may not always be valid in ProLex, indicating future research opportunities in using LLMs for generating semantically correct, collocationally appropriate English lexical substitutions.


\begin{figure}[h]
\centering
\includegraphics[width=0.35\textwidth]{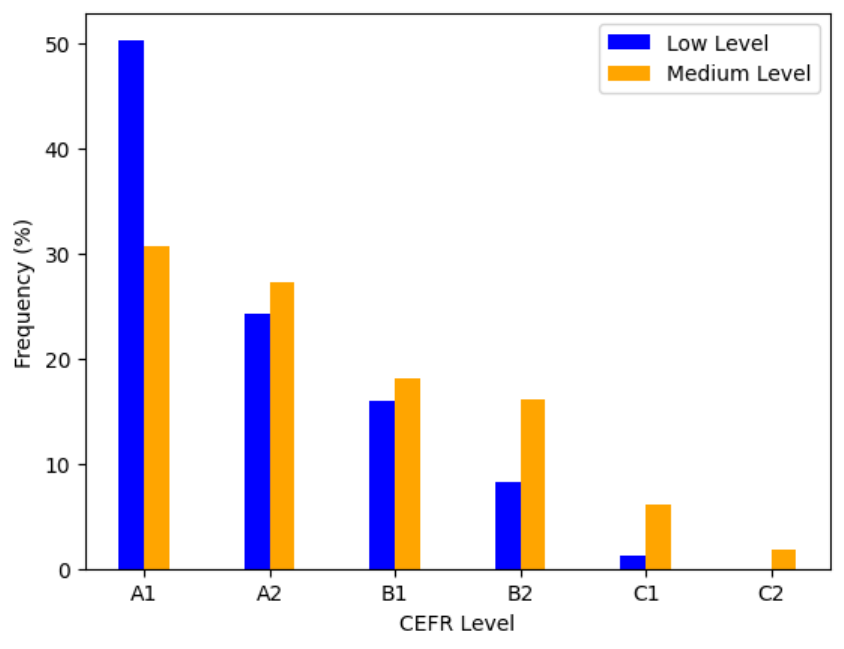}
\caption{Distribution of CEFR levels of target words in low and medium sentences in ProLex.}
\label{fig:t_words_cefr}
\end{figure}

\begin{figure*}[h]
\centering
\subfigure[Acceptable substitutes $w^a$.]{
\begin{minipage}[t]{0.5\linewidth}
\centering
\includegraphics[width=2.3in]{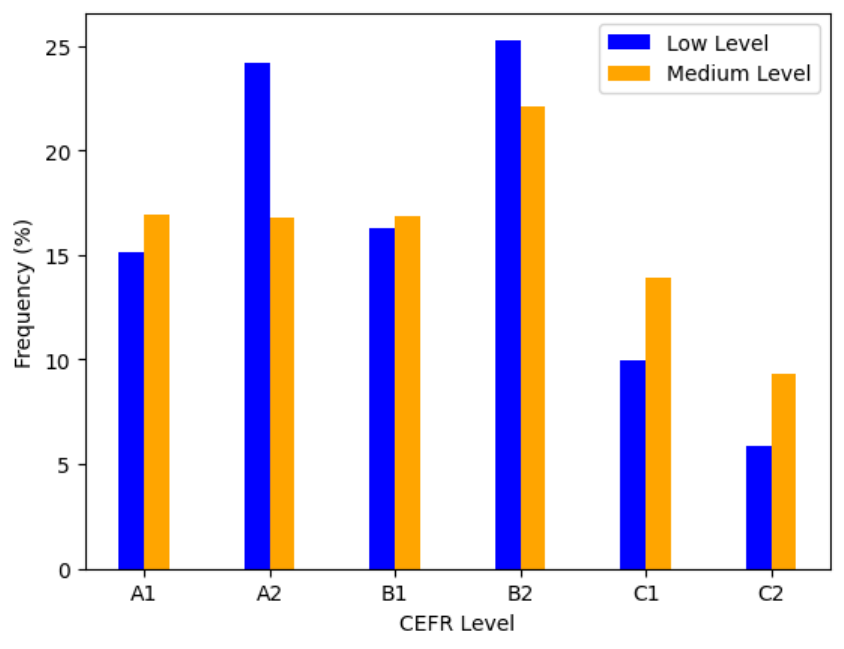}
\label{fig:subs_cefr_1}
\end{minipage}%
}%
\subfigure[Proficiency-oriented substitutes $w^a_p$.]{
\begin{minipage}[t]{0.5\linewidth}
\centering
\includegraphics[width=2.3in]{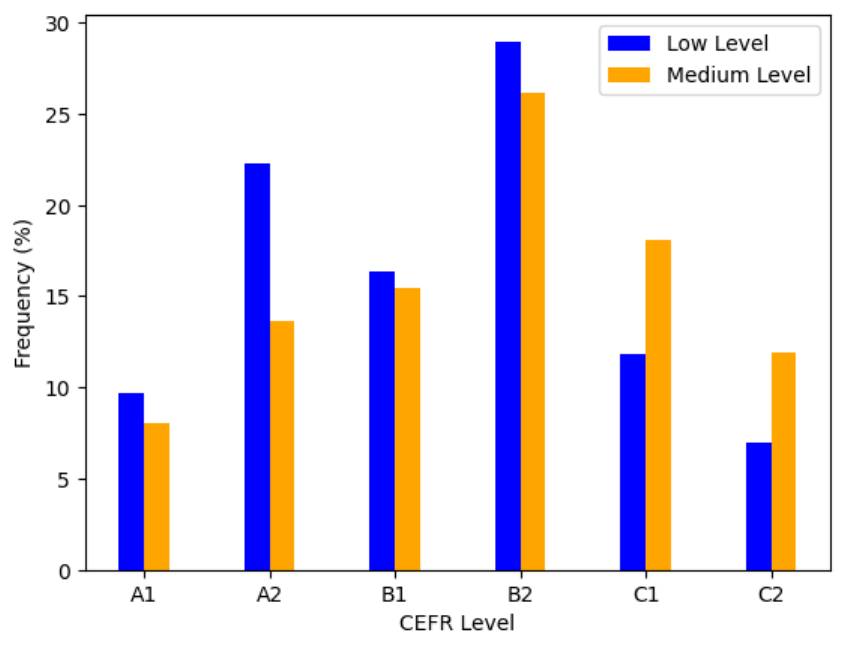}
\label{fig:subs_cefr_2}
\end{minipage}%
}%
\caption{Distribution of CEFR levels for substitutes in $w^a$ and substitutes in $w^a_p$ in ProLex. In low-level sentences, more than 65\% of the proficiency-oriented substitutes are from B1 level or higher; similarly, in medium-level sentences, over 75\% of these substitutes are sourced from B1 level or above. }
\label{fig:subs_cefr}
\end{figure*}

We further process the annotations to compose the final quadruplets $(w, s, w^a, w^a_p)$ in ProLex, where $w^a$ denotes a list of acceptable substitutes for $w$, and $w^a_p$ denotes the list of acceptable substitutes after filtering based on CEFR levels. We exclude the target words that have zero acceptable substitutes. Table \ref{tab:stat_test} provides an overview of the statistics for the resulting data, grouped by low and medium-level proficiency. In total, there are $680$ word-sentence $(w, s)$ pairs in ProLex with at least one acceptable substitute.\footnote{The Part-Of-Speech distribution for the target word $w$ in ProLex is shown in Table \ref{tab:stat_test_tword_pos}.} On average, for each $(w, s)$ pair, there are $2.9$ acceptable substitutes $w^a$ and $2.3$ proficiency-oriented substitutes $w^a_p$. Furthermore, we provide the distribution of CEFR levels of the target words in both low and medium sentences, and the result is shown in Figure \ref{fig:t_words_cefr}. As evident, a significant proportion of the chosen target words are from A1 and A2 levels, comprising 75\% in low-level sentences and 55\% in medium-level sentences, respectively. Moreover, in Figure \ref{fig:subs_cefr}, we also present the CEFR level distributions for acceptable substitutes $w^a$ and proficiency-oriented substitutes $w^a_p$. The distribution resonates closely with our research objective of ProLex, which aims to propose acceptable substitutions of high-proficiency level words for commonly used low-proficiency level words in L2 English sentences.



\section{Language Proficiency-oriented Lexical Substitution}

In this section, we outline recommended evaluation practices for ProLex, propose models that can automatically perform the task on ProLex, and assess the performance of out-of-the-box LLMs and top lexical substitution systems. 


\subsection{Evaluation Settings}

For the task of lexical substitution, there are two mainstream evaluation settings: generative setting \cite{mccarthy2007semeval} and ranking setting \cite{thater2010contextualizing}. In the generative context, systems produce a sequence of potential substitutes, and there are no limits on the number of candidate substitutes they can generate. In the ranking scenario, systems receive the entire set of substitute options provided by the benchmark. The task then is to order these substitutes according to their appropriateness. In this work, we mainly focus on the generative setting and defer the ranking setting to future research.


\begin{table*}[]
\centering
\begin{adjustbox}{max width=0.9\textwidth}
\begin{tabular}{ll}
\toprule
Target word ($w$)                  & promotion (B2)                                                                                               \\
Sentence ($s$)                     & This **promotion** has a beautiful and effective visual part, but they miss the real point: the product. \\
Acceptable ($w^a$)           & advertising (A2), marketing (B1), publicity (B2), campaign (B1), advertisement (A2)                                               \\
Proficiency-oriented ($w^a_p$) & \textcolor{blue}{publicity} (B2)                                                                                                \\ \midrule
\textit{GPT-4 (32-shot)}                          & \textbf{advertising}, \textbf{marketing}, \textcolor{blue}{\textbf{publicity}}, hype, announcement         \\
\textit{ChatGPT (32-shot)}                        & \textbf{advertising}, \textbf{marketing}, \textcolor{blue}{\textbf{publicity}}, \textbf{campaign}, endorsement      \\
\textit{Vicuna-1.5-13B-$D_L$}                & advancement, \textcolor{blue}{\textbf{publicity}}, \textbf{marketing}, endorsement                                                           \\
\textit{Llama2-13B-$D_{LS}$}                & advancement, progression, elevation                                                          \\
\textit{Para-LS}                & \textbf{advertising}, \textbf{advertisement}, \textbf{marketing}, \textbf{campaign}, \textcolor{blue}{\textbf{publicity}}, ad, presentation, show, event, production                                                           \\

\bottomrule
\end{tabular}
\end{adjustbox}
\caption{Example data point and predictions from the top systems. In the outputs, the acceptable substitutes generated from each systems are bolded, while proficiency-oriented ones are both bolded and blue.}
\label{tab:qua_comp}
\end{table*}

\subsection{Evaluation Metrics}
We evaluate the system performance in terms of its substitute \textit{appropriateness} and \textit{proficiency}. For assessing \textit{appropriateness}, we compare the system's predictions with the acceptable substitutes, denoted as $w^a$; similarly, for evaluating \textit{proficiency}, we contrast the system's predictions with proficiency-oriented substitutes, represented as $w^a_p$. Inspired by \citet{lee2021swords}, we consider the evaluation metrics that measure the quality and coverage of the predicted substitutes from a system. Specifically, for \textit{appropriateness}, we compute precision ($P^k$), recall ($R^k$) and F-score ($F^k$)\footnote{$F^k$ is calculated as the harmonic mean of $P^k$ and $R^k$.} at $k$:

\begin{gather*}
P^k = \frac{\# \text{ acceptable subs } w^a \text{ in system top-}k}{\#\text{ substitutes in system top-}k} \\
R^k = \frac{\# \text{ acceptable subs } w^a \text{ in system top-}k}{\min(k, \# \text{ acceptable subs } w^a)}
\end{gather*}

Similarly, we evaluate \textit{proficiency} against the list of proficiency-oriented substitutes $w^a_p$. $P^k_{p}$ and $R^k_{p}$ represent the precision and recall of system outputs against this smaller candidate list, with $F^k_{p}$ as their harmonic mean. Also, we follow previous work \cite{mccarthy2007semeval,lee2021swords} to mainly examine performance for $k = 10$, and implement a \textit{soft} evaluation setting, where all substitutes generated by a system are lemmatized before comparison. In addition, given that ProLex considers grammatically correct substitutes, we also apply a \textit{hard} evaluation setting, where predictions should \textit{exactly match} the reference without lemmatization.

\begin{table*}[]
\centering
\begin{adjustbox}{max width=0.95\textwidth}
\begin{tabular}{llcc|llll|llll|ccll}
\toprule
\multicolumn{2}{l}{\multirow{3}{*}{\textbf{Models}}} & \multicolumn{1}{l}{\multirow{3}{*}{\textit{BERT-LS}}} & \multicolumn{1}{l|}{\multirow{3}{*}{\textit{Para-LS}}} & \multicolumn{4}{c|}{\textit{Llama-2}}                                                                                & \multicolumn{4}{c|}{\textit{Vicuna-1.5}}                                                                             & \multicolumn{2}{c}{\multirow{2}{*}{\textit{ChatGPT}}} & \multicolumn{2}{c}{\multirow{2}{*}{\textit{GPT-4}}}        \\ \cline{5-12}
\multicolumn{2}{l}{}                                 & \multicolumn{1}{l}{}                         & \multicolumn{1}{l|}{}                         & \multicolumn{2}{c}{7B}                               & \multicolumn{2}{c|}{13B}                             & \multicolumn{2}{c}{7B}                               & \multicolumn{2}{c|}{13B}                             & \multicolumn{2}{c}{}                         & \multicolumn{2}{c}{}                              \\ \cline{5-16} 
\multicolumn{2}{l}{}                                 & \multicolumn{1}{l}{}                         & \multicolumn{1}{l|}{}                         & \multicolumn{1}{c}{$D_S$} & \multicolumn{1}{c|}{$D_L$} & \multicolumn{1}{c}{$D_S$} & \multicolumn{1}{c|}{$D_L$} & \multicolumn{1}{c}{$D_S$} & \multicolumn{1}{c|}{$D_L$} & \multicolumn{1}{c}{$D_S$} & \multicolumn{1}{c|}{$D_L$} & $zero$        & \multicolumn{1}{c|}{$32$}        & \multicolumn{1}{c}{$zero$} & \multicolumn{1}{c}{$32$} \\ \midrule
\multirow{2}{*}{\textbf{Soft}}         & $F^{10}$        & 25.2                                         & 30.2                                          & 44.1                     & 48.6                      & 44.4                     & 51.1                      & 43.5                     & 48.0                      & 46.2                     & 52.1                      & 50.9        & 51.5                           & 50.5                     & \textbf{54.7}                   \\
                                       & $F^{10}_p$        & 20.3                                         & 25.3                                          & 39.1                     & 43.2                      & 38.6                     & 46.2                      & 38.6                     & 43.1                      & 40.7                     & 46.8                      & 45.2        & 45.9                           & 44.6                     & \textbf{48.8}                   \\ \midrule
\multirow{2}{*}{\textbf{Hard}}         & $F^{10}$       & 20.0                                         & 20.2                                          & 31.2                     & 46.5                      & 27.1                     & 49.0                      & 31.4                     & 45.6                      & 32.2                     & 49.6                      & 47.8        & 48.3                           & 48.8                     & \textbf{52.7}                   \\
                                       & $F^{10}_p$      & 15.8                                         & 17.2                                          & 27.8                     & 41.2                      & 23.9                     & 44.1                      & 27.7                     & 40.9                      & 28.6                     & 44.5                      & 42.5        & 43.0                           & 43.3                     & \textbf{46.8}                \\ \bottomrule
\end{tabular}
\end{adjustbox}
\caption{Evaluation of systems on ProLex under \textit{soft} and \textit{hard} settings. We fine-tuned \textit{Vicuna} and \textit{Llama-2} model variants on both $D_L$, the synthesized data generated by LLM, and $D_S$, the filtered dataset based on Swords \cite{lee2021swords}. We also conducted zero-shot (i.e. $zero$) and in-context learning (i.e. $32$ shots) with LLMs.}
\label{tab:eval_results}
\end{table*}

\begin{table}[]
\centering
\begin{adjustbox}{max width=0.75\linewidth}
\begin{tabular}{llcccc}
\toprule
\multicolumn{2}{l}{\multirow{2}{*}{\textbf{Models}}} & \multicolumn{2}{c}{\textit{Vicuna-1.5}} & \multicolumn{2}{c}{\textit{Llama-2}} \\ \cmidrule{3-6} 
\multicolumn{2}{l}{}                                 & 7B   & \multicolumn{1}{c}{13B} & 7B       & 13B              \\ \midrule
\multirow{2}{*}{\textbf{Soft}}         & $F^{10}$       & 49.9 & 52.4                     & 49.7     & \textbf{54.2}    \\
                                       & $F^{10}_p$       & 45.0 & 47.0                     & 44.6     & \textbf{48.4}    \\ \midrule
\multirow{2}{*}{\textbf{Hard}}         & $F^{10}$        & 48.1 & 50.5                     & 48.0     & \textbf{51.6}    \\
                                       & $F^{10}_p$        & 43.0 & 45.2                     & 42.8     & \textbf{45.9}    \\ \bottomrule
\end{tabular}
\end{adjustbox}
\caption{Evaluation results on ProLex after fine-tuning \textit{Vicuna-1.5} and \textit{Llama-2} on the aggregate dataset $D_{LS}$. }
\label{tab:combined_train}
\end{table}

\subsection{Baselines}

We evaluate systems in the following settings: 1) existing top lexical substitution systems, 2) instruction-tuning language models on task-specific synthetic data, and 3) zero-shot prompting and in-context learning with LLMs. 

\paragraph{Existing Top Lexical Substitution Systems}
Past lexical substitution systems were proposed and evaluated on three widely-known benchmarks: LS07 \cite{mccarthy2007semeval}, CoInCo \cite{kremer2014substitutes} and Swords \cite{lee2021swords}. We evaluate the following two representative systems on ProLex:
\begin{itemize}
    \item \textit{BERT-LS} \cite{zhou2019bert}: a BERT-based lexical substitution system. It once reached SOTA results on LS07 and CoInCo dataset.
    \item \textit{Para-LS} \cite{qiang2023parals}: a paraphraser-based system, achieving SOTA performance on all three benchmarks: LS07, CoInCo, and Swords.
\end{itemize}

\paragraph{Instruction Tuning with Task-specific Synthetic Data} As instruction tuning has proliferated in building powerful instruction following models \cite{ouyang2022training,bai2022training,openai_gpt4, chiang2023vicuna}, we experiment with two instruction-tuned large-scale language models, namely \textit{Vicuna}\footnote{We used Vicuna v1.5 provided by \cite{zheng2023judging}.} \cite{chiang2023vicuna} and \textit{Llama-2} \cite{touvron2023llama}. Note that we mainly focus on the smaller variants: \textit{Vicuna} 7B/13B and \textit{Llama-2} 7B/13B. Since there is no existing training data for our language proficiency-oriented lexical substitution task, we adopt two approaches to synthesize the data: 1) directly generate data with GPT-4 through prompting, and 2) modify the existing lexical substitution benchmark, Swords. The prompts we employed to synthesize the data with GPT-4 are shown in section \ref{syn_data}. As for the second approach, we considered only the acceptable substitutes (i.e. score greater than 50\%) from the Swords dataset and filtered the substitutes based on their CEFR levels, by only keeping the ones that have equal or higher proficiency levels compared to the target word. As a result, we denote the training data synthesized with GPT-4 as $D_L$, the data modified from Swords as $D_S$, and the aggregate of these two datasets as $D_{LS}$.

\paragraph{Zero-shot and In-context Learning with LLMs} Leveraging the potent zero-shot capabilities of Large Language Models (LLMs), we evaluate their performance on ProLex by prompting two widely recognized LLMs, \textit{GPT-4} and \textit{ChatGPT},\footnote{Note that we used the version \texttt{gpt-3.5-turbo-1106}} to generate potential substitution candidates. Moreover, we also perform in-context learning with examples selected from $D_L$.\footnote{We randomly select $32$ examples from $D_L$ for in-context learning with LLMs} See appendix \ref{prompt_gpt_zero} for more details.

\subsection{Results and Analysis}
Table \ref{tab:eval_results} demonstrates the results of our experiments on ProLex. We take the top 10 ($k=10$) predictions from each system to evaluate. In general, the \textit{appropriateness} scores $F^{10}$ for both \textit{soft} and \textit{hard} settings are greater than the \textit{proficiency} scores $F^{10}_p$, indicating the challenges of generating proficiency-oriented substitutes. Additionally, all systems tend to achieve higher scores in \textit{soft} setting, given that all predictions are lemmatized before comparison. This indicates that these systems might produce substitutes that are grammatically incorrect yet suitable and demonstrate advanced language proficiency.


Overall, in-context learning with \textit{GPT-4} achieved the best results across all systems in all evaluation settings. Note that the 32 in-context examples were randomly selected from our task-specific synthetic dataset $D_L$. This underscores the effectiveness of our data synthesis method, which can be applied in situations with limited data resources. Using $D_L$, \textit{Vicuna-1.5-13B} achieved the best performance among all fine-tuned model variants on single dataset and previous top-performing lexical substitution systems, even surpassing \textit{ChatGPT} on both zero-shot and in-context learning scenarios, and \textit{GPT-4} on zero-shot setting.


In addition, in our instruction-tuning setup, training with dataset $D_S$ resulted in lower performance compared to training with the synthetic dataset $D_L$. This occurred because $D_S$ is derived from filtering Swords, which comprises context sentences not authored by L2 English learners, thus leading to a distribution shift. On the other hand, all context sentences in dataset $D_L$ originate from the TOEFL-11 corpus, the same source used to develop ProLex. We also conducted fine-tuning experiments using the aggregate dataset $D_{LS}$. Table \ref{tab:combined_train} demonstrates that combining $D_S$ and $D_L$ enhanced the performance of all fine-tuned models. Notably, \textit{Llama-2-13B} achieved scores nearly comparable to the best result of \textit{GPT-4}.

As for previous lexical substitution systems, \textit{Para-LS} consistently surpasses \textit{BERT-LS } but lags far behind all other larger-scale systems. This highlights the limitations of these earlier methods and points towards a promising research direction in developing more efficient systems that can perform better in generating language proficiency-oriented substitutions.

For error analysis, Table \ref{tab:qua_comp} shows an example and the outputs generated by top systems. \textit{GPT-4} and \textit{ChatGPT} produce words that have different semantics from the target (e.g. "hype" for **promotion**). Our fine-tuned models produce substitutes covering multiple connotations of the target (e.g. "advancement" and "marketing"), leading to lower precision. \textit{Para-LS} generates words that have a more general meaning than the target (e.g. "event" and "show").

\section{Conclusion and Future Work}

We propose a new task \textit{language proficiency-oriented lexical substitution} to improve the vocabulary diversity and writing proficiency of L2 English learners. We introduce ProLex, a novel benchmark designed to assess systems' ability to generate \textit{appropriate} and \textit{language proficiency-oriented} substitutes for given target words in the context. Besides, we also fine-tuned open-source language models on task-specific synthetic data, achieving results that are on par with, or better than GPT-4 and ChatGPT. In future, we will expand ProLex by leveraging larger L2 English writing corpora and incorporating more comprehensive sets of substitutes.

\section{Limitations}

\paragraph{Benchmark Size and Coverage}
Compared to previous Lexical Substitution benchmarks \cite{mccarthy2007semeval, kremer2014substitutes, lee2021swords}, the size of ProLex is relatively small. However, considering that ProLex draws its data from the TOEFL-11 corpus, it could be significantly extended by incorporating sentences from additional sources authored by L2 speakers.

In addition, the systems may produce valid substitutes that are not present in ProLex during evaluation, indicating that ProLex has limitations in coverage. By employing the annotation scheme described in Section \ref{sec:ann_main}, we can iteratively update and enhance ProLex with additional gold standard substitutes, thereby expanding its coverage in future versions.




\paragraph{Limits of CEFR Checker}
The CEFR Checker \cite{cathovenai2023} we used in this work is capable of assigning CEFR levels only at the word level. However, we noticed that in some cases, phrases could also serve as appropriate substitutes. Hence, extending the CEFR Checker's functionality to include CEFR-level assignments at both the word and phrase levels would offer a more holistic evaluation in future versions of ProLex.

\paragraph{Phrases and Multi-word Substitutions}
In this work, we posit that the integration of multi-word expressions, such as idiomatic phrases, can enhance English language proficiency \cite{thyab2016necessity,yunus2021influence,al2016figurative}. Therefore, we treat all acceptable phrasal substitutes as valid substitutes for demonstrating better language proficiency compared to the single target word. In future research, we intend to investigate the methods for assessing proficiency levels in the use of phrases and multi-word expressions.


\section{Ethics Statement}
\paragraph{Reproducibility}
In this work, our data creation process utilized GPT-4. We also used ChatGPT for evaluation purposes. Although they are not open-sourced language models, to facilitate the reproducibility of our results, we demonstrated all of prompts used in our paper. In addition, all the other models used in this research, are publicly available in peer-reviewed articles and referenced in this paper. All datasets, including our synthetic fine-tuning dataset and all annotated test data, will be released. 

\paragraph{Biases} Our models are built over \textit{Vicuna} and \textit{Llama-2}. In this work, we did not explicitly handle
any bias that exists in the two pre-trained language models.

\paragraph{Human Annotators}

Both annotators were recruited from the doctoral programs in the linguistics department, and they specialize in English language teaching and education. They were paid at a rate of \$13 per hour. To ensure the privacy and anonymity of all contributors, no personal or demographic information was collected.

\section{Acknowledgement}
We would like to thank Qingyang Wu, Kun Qian, Siyan Li, Matthew Toles, Yu Li, and Xiao Yu for their valuable discussions and suggestions on the paper. We also want to express our gratitude to Cathoven AI for providing free access to their CEFR checker APIs. In addition, we thank our expert annotators for their time and contributions to the completeness of the benchmark.

\bibliography{anthology,custom}
\bibliographystyle{acl_natbib}

\clearpage

\appendix

\section{Detailed ProLex Annotation Scheme}
\label{sec:ann_scheme}
In this section, we present the detailed annotation scheme for ProLex. In the annotation process, each annotator is presented with $2,500$ $(w, s, w')$ triplets. For each $(w, s, w')$, we ask the annotators to judge whether $w'$ is an acceptable substitute for $w$ in $s$. In particular, they will choose among three labels: $1$ (acceptable), $-1$ (unacceptable), and $0$ (uncertain). The general guidelines are shown in Table \ref{tab:ann_rules}. A candidate substitute $w'$ is labeled as $1$ only if it satisfies all criteria in the table. It is labeled as $-1$ once it violates any one of them. We also provide label $0$ to exclude the cases when either the target word $w$ or context sentence $s$ is extremely hard to understand as they are written by beginner L2 English learners. Besides the general guidelines, we also provide a detailed specification along with some examples for all three labels. See Table \ref{tab:ann_examples} for more illustration examples. The detailed specification is as follows:
\begin{itemize}
    \item To determine if a substitute is a common collocation or expression in actual English use, annotators should refer to COCA to search for certain expressions:
    \begin{itemize}
        \item Mark $1$ if the frequency of the expression is greater than 5
        \item Mark $-1$ otherwise
    \end{itemize}
    \item Mark $1$ if the substitute is an \textbf{entailment} from the sense of the target word.
    \item Mark $1$ if the substitute can be used \textbf{figuratively} and also convey the same semantics as the target word does.
    \item Mark $0$ if the substitute is a hyponym of the target word.
    \item A sentence can contain some grammar errors (e.g. spelling errors). If it is written in a way that it is hard for you to understand, mark it $0$; otherwise, if you can understand, mark the substitute as $1$ or $-1$.
    \item Mark $1$ if the substitute is good even though with incorrect determiner like “a” vs “an”.
    \item Mark $-1$ if the substitute is good but not quite grammatically correct.
    \item Grammar of a substitute matters, please keep a high bar for making a substitute as $1$.
    \item Mark $1$ if the target word itself appears as a substitute.
    \item Mark $-1$ if the target word is a proper noun or part of a fixed expression/phrase.
    \item Mark $0$ if you need more contexts of the original sentence to help you judge the acceptability.
\end{itemize}

\section{Generate Candidate Substitutions from GPT-4}
\label{sec:gpt4_prompts}

We generate the candidate substitutes for human annotation by prompting GPT-4. The prompt we used during generation is demonstrated in Table \ref{tab:gpt4_prompt_gen}. In particular, we selected five instances from Swords \cite{lee2021swords} as the in-context examples to guide the generation process. In these examples, the target words consist exclusively of content words — nouns, verbs, adjectives, and adverbs — which represent the primary focus of ProLex.

\section{Dataset Statistics}
\label{sec:stat_tword_w}

In this section, we illustrate the statistics of all target words and context sentences extracted from Section \ref{sec:data_col}. The overall statistics are shown in Table \ref{tab:stat_tword_sel} below. In general, there are $2,531$ unique $(w, s)$ pairs combing sentences from low and medium-level essays. Note that on average, sentences from medium-level essays are longer than the ones from low-level essays. Furthermore, nouns comprise the majority of the target words, accounting for 57\% of all selected words. In contrast, verbs, adjectives, and adverbs make up 23\%, 16\%, and 4\% of the selection, respectively.

\begin{table*}
\begin{adjustbox}{max width=\textwidth}
    \centering
    \begin{tabular}{c|c|c|c|c|c|c|c} \hline 
         \textbf{Essay Proficiency}&  \textbf{\# of ($w$, $s$)}& \textbf{ \# of unique $s$}& \textbf{Avg $s$ length} & \textbf{\# of nouns}&  \textbf{\# of verbs}&  \textbf{\# of adj}& \textbf{\# of adv}\\ \hline 
         low&  630&  609& 19& 345&  169&  94& 22\\ 
         medium&  1901&  1824& 64&  1090&  415&  320& 76\\ 
         all&  2531&  2433& 21&  1435&  584&  414& 98\\ \hline
    \end{tabular}
\end{adjustbox}
\caption{Statistics of all target words $w$ and context sentences $s$. ($w$, $s$) indicates a pair of target words and contexts. "all" means the combinations of both "low" and "medium" essays.}
\label{tab:stat_tword_sel}
\end{table*}

Note that we sample $1,000$ $(w, s)$ pairs from the global dataset, generate candidate substitutes from GPT-4, and ask human annotators to judge the contextual appropriateness of these candidates. After filtering the ones labeled with $0$, we end up with $680$ $(w, s)$ pairs in the final ProLex dataset. We also present the statistics of these $(w, s)$ pairs in ProLex in Table \ref{tab:stat_test_tword_pos}. Among all $680$ pairs, 52\% of them are nouns, 25\% are verbs, 18\% are adjectives, and 5\% are adverbs, closely mirroring the distribution found in the overall dataset described above.


\begin{table*}
\centering
\begin{adjustbox}{max width=\textwidth}
    \centering
    \begin{tabular}{c|c|c|c|c|c} \hline 
         \textbf{Essay Proficiency}&  \textbf{\# of ($w$, $s$)}& \textbf{\# of nouns}&  \textbf{\# of verbs}&  \textbf{\# of adj}& \textbf{\# of adv}\\ \hline 
         low&  169& 83&  50&  27& 9\\ 
         medium&  511&   268&  122&  95& 26\\ 
         all&  680&   351&  172&  122& 35\\ \hline
    \end{tabular}
\end{adjustbox}
\caption{Statistics of target words $w$ in ProLex in terms of the part-of-speech tags.}
\label{tab:stat_test_tword_pos}
\end{table*}

\section{Grammar Error Correction on TOEFL-11}
\label{app:grammar_error}

Because the sentences in ProLex are extracted from low and medium-level essays from TOEFL-11 corpus, these sentences frequently contain basic grammar errors that obscure the intended meaning of the context. To alleviate this problem, we first follow \citet{rothe2021simple} and tested a T5 base model trained on cLang-8 on the CoNLL-14 test set. Finding its performance unsatisfactory, we further fine-tuned the model on ErAConD \cite{yuan2022eracond} which slightly improved it. At last, we chose to fine-tune a GPT-2 model \cite{radford2019language} on the proposed cLang-8 dataset, which provided a satisfactory $F_{0.5}$-score as shown in table \ref{tab:gec_scores}, allowing us to correct most grammatical errors in the TOEFL-11 essays.

\begin{table*}
\centering
\begin{adjustbox}{max width=\textwidth}
    \centering
    \begin{tabular}{c|c|c|c|c} \toprule 
         \textbf{Model}&  \textbf{Fine-tuning Data}& \textbf{Precision} & \textbf{Recall} & \textbf{$F_{0.5}$-score}  \\ \midrule 
         T5-base &  cLang-8 & 59.7 & 26.1 & 47.5 \\ 
         T5-base &  cLang-8 + ErAConD & 60.6 & 36.3 & 53.4 \\ 
         GPT2-large & cLang-8 & \textbf{66.8} & \textbf{48.1} & \textbf{61.9} \\ \bottomrule
    \end{tabular}
\end{adjustbox}
\caption{Performance of different GEC systems on the CoNLL-14 Shared Task test dataset, as measured by the official $M^2$Scorer \cite{dahlmeier2012better}.}
\label{tab:gec_scores}
\end{table*}

In Table \ref{tab:gec_examples}, we provide some examples to show the original sentences from TOEFL-11 and the corresponding corrected sentences output from our grammar model.

\begin{table*}[]
\begin{adjustbox}{max width=\textwidth}
\centering
\begin{tabular}{ll}
\toprule
\multicolumn{1}{c}{\textbf{Original Sentence}}                                                                                                                                  & \multicolumn{1}{c}{\textbf{Corrected Sentence}}                                                                                                                                       \\ \midrule
\begin{tabular}[c]{@{}l@{}}Meanwhile when you have a tour guide, that means you are safer especially \\ if \textcolor{red}{he} from the area which you will \textcolor{red}{go} that will be better.\end{tabular} & \begin{tabular}[c]{@{}l@{}}Meanwhile when you have a tour guide, that means you are safer especially \\ if \hl{\textbf{he is}} from the area which you will \hl{\textbf{go to}} that will be better.\end{tabular} \\ \midrule
\begin{tabular}[c]{@{}l@{}}but if you are a man who isn't \textcolor{red}{open} new ideas and you \textcolor{red}{are't} produce new \\ things you are not \textcolor{red}{be} a good employee or a good boss.\end{tabular}       & \begin{tabular}[c]{@{}l@{}}but if you are a man who isn't \hl{\textbf{open to}} new ideas and you \hl{\textbf{don't}} produce new \\ things you are \hl{\textbf{not a}} good employee or a good boss.\end{tabular}              \\ \midrule
There are two reasons for this \textcolor{red}{statements}: time and \textcolor{red}{knowridge}.                                                                                                                  & There are two reasons for this \hl{\textbf{statement}}: time and \hl{\textbf{knowledge}}.                                                                                                                         \\ \bottomrule
\end{tabular}
\end{adjustbox}
\caption{Examples of basic grammar error correction for sentences selected from TOEFL-11. The erroneous parts are marked in red, and the corresponding corrected portions are highlighted in yellow.}
\label{tab:gec_examples}
\end{table*}

\begin{figure*}[h]
\centering
\subfigure[Target word in $D_S$.]{
\begin{minipage}[t]{0.5\linewidth}
\centering
\includegraphics[width=3in]{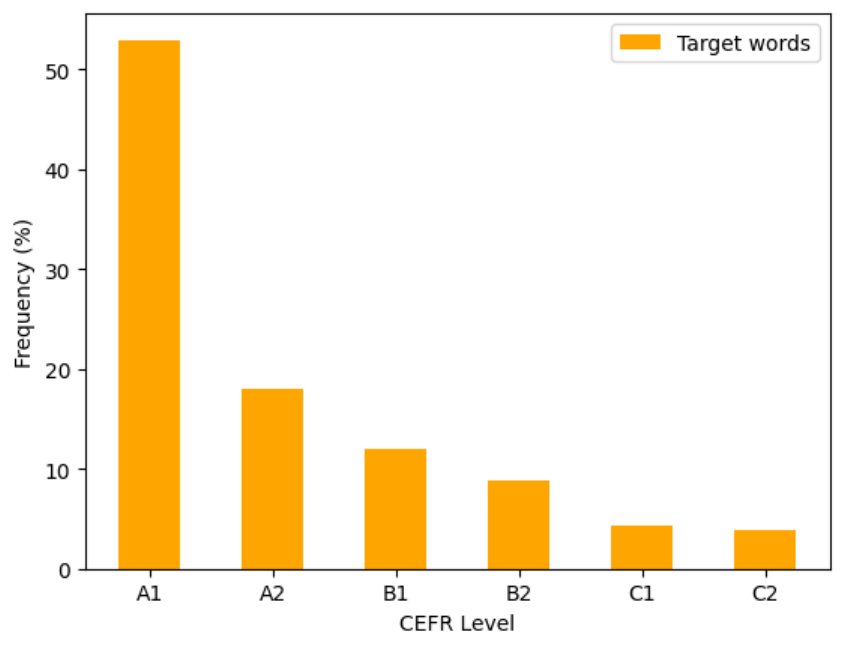}
\label{fig:swords_subs_cefr_1}
\end{minipage}%
}%
\subfigure[Proficiency-oriented substitutes in $D_S$.]{
\begin{minipage}[t]{0.5\linewidth}
\centering
\includegraphics[width=3in]{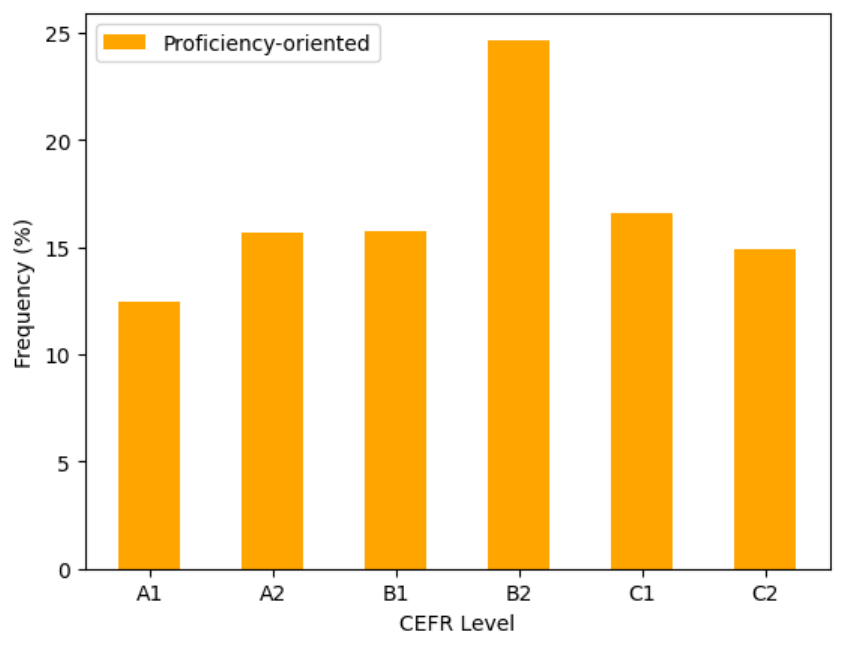}
\label{fig:swords_subs_cefr_2}
\end{minipage}%
}%
\caption{Distribution of CEFR levels for target words and proficiency-oriented substitutions in $D_S$.}
\label{fig:swords_filtered_cefr}
\end{figure*}


\section{Details of Baseline Experiments}

The details of the baseline experiments are presented in this section. We measure the performance of the following baseline systems: 1) zero-shot LLMs prompting, 2) instruction tuning with task-specific synthesized data, and 3) current state-of-the-art lexical substitution systems. In the following, the setups for each of these baselines are illustrated in great detail.

\subsection{Zero-shot and In-context Learning with GPT-4 and ChatGPT}
\label{prompt_gpt_zero}

We evaluate the zero-shot performance of GPT-4 and ChatGPT on ProLex. The prompt we used in the experiment is shown in Table \ref{tab:zero_prompt}. Specifically, we used the model provided by OpenAI platform, namely \texttt{gpt-4} and \texttt{gpt-3.5-turbo-1106}, to evaluate the performance of GPT-4 and ChatGPT, respectively. For each $(w, s)$ pair, GPT-4 and ChatGPT outputs a list of substitutes that are separated by commas. In our evaluation, we take the top-{$k$} $(k=10)$ substitutes from the outputs and compare them with our labels in ProLex. Similarly, we also measure the performance of GPT-4 and ChatGPT under in-context learning settings. We randomly select $32$ examples from the synthetic dataset $D_L$ and use these as in-context examples for both GPT-4 and ChatGPT. The prompt we used for this experiment is shown in Table \ref{tab:icl_prompt}. 

\subsection{Instruction tuning experiments}

In our instruction tuning experiments, we conducted evaluations on the 7B and 13B variants of \textit{Vicuna} and \textit{Llama-2}. To synthesize task-specific training data, we applied two approaches: 1) generate data with GPT-4 and 2) modify the existing lexical substitution benchmark Swords. In the following, 
we will present detailed descriptions of the two approaches, along with the specific experimental settings used in the fine-tuning process.

\subsubsection{Synthesizing task-specific training data}
\label{syn_data}

\paragraph{Generate data from GPT-4}
We start by selecting the context sentences from the TOEFL-11 dataset. Concretely, we randomly select $500$ sentences from each proficiency level,\footnote{Note that there are three proficiency levels in the TOEFL-11 dataset, namely high, medium and low.} rendering a total of $1,500$ sentences. We also make sure that the selected sentences are not in ProLex. Similar to section \ref{sec:data_col}, we perform grammar error correction to the sentences extracted from low and medium-level essays, while sentences from high-level essays are retained in their original form. Subsequently, for each sentence, we prompt GPT-4 to select a target word based on a randomly pre-defined part-of-speech tag and generate the proficiency-oriented substitutes. The complete prompt for this process is shown in Table \ref{tab:gpt4_syn}. Specifically, we provide five in-context examples to guide the generation process. As a result, we take the final substitutes generated for each sentence as the proficiency-oriented substitutes. The resulting synthesized data is post-processed to fine-tune \textit{Vicuna} 7B/13B and \textit{Llama-2} 7B/13B models. In total, after post-processing,\footnote{The post-processing involves dropping null values and duplicated rows.} the synthesized dataset contains $1,383$ unique $(w, s)$ pairs, with $6,375$ candidate substitutes and $4,982$ final substitutes. The part-of-speech tags for all selected target words are uniformly distributed, with $361$ nouns, $335$ verbs, $357$ adjectives, and $330$ adverbs.
 
\paragraph{Modify existing benchmark}

Another way to synthesize training data is to modify the dataset provided by the Swords benchmark, filtering out lower proficiency level substitutes. Concretely, we only consider acceptable substitutes (i.e. score greater than 50\%) in Swords, and perform filtering based on their CEFR levels.


The Swords dataset, with 1132 target word and context pairs, have an average of $60.7$ substitutes per target, of which $4.1$ are acceptable on average. After filtering out substitutes with a lower CEFR proficiency level than the target word, the dataset now contains an average of $3$ substitutes per target. The average CEFR level of targets is A2, whereas the average level of substitutes after filtering is B1, indicating improvements on proficiency-level. The CEFR level distributions of the target words and proficiency-oriented substitutes from the filtered dataset $D_S$ are shown in Figure \ref{fig:swords_filtered_cefr}.



\subsubsection{Experiment setup}

We established our training pipeline based on the platform developed by \cite{zheng2023judging}. Concretely, for the 7B models, we fine-tuned both \textit{Vicuna} and \textit{Llama-2} for a maximum of 10 epochs each, using a single NVIDIA A100-80G GPU. For the 13B models, each variant was trained for up to 5 epochs on two NVIDIA A100-80G GPUs. In all cases, we configured the training batch size per device to $1$ and established the initial learning rate at \texttt{1e-5}, employing a linear learning rate scheduler. The best checkpoints were selected based on the performance on a separate development set of ProLex.

\begin{table*}[]
\centering
\begin{adjustbox}{max width=\textwidth}
\begin{tabular}{cl}
\toprule
\textbf{Label} & \multicolumn{1}{c}{\textbf{Criteria}}                                                                                                                                                                                                                                              \\ \midrule
1              & \begin{tabular}[c]{@{}l@{}}$\cdot$ doesn’t change the meaning and semantics of the sentence, \textbf{and}\\ $\cdot$ is a common collocation or expression in actual English use, \textbf{and}\\ $\cdot$ in general, matches at least one connotation of the target word in the \\ context, \textbf{and}\\ $\cdot$ grammatically correct\end{tabular} \\ \midrule
-1             & \begin{tabular}[c]{@{}l@{}}$\cdot$ changes the meaning and semantics of the sentence, \textbf{or}	\\ $\cdot$ is not a common collocation or expression in actual English use \textbf{or}	\\ $\cdot$ doesn’t match any connotation of the target word in the context \textbf{or}	\\ $\cdot$ is grammatically incorrect\end{tabular}              \\ \midrule
0              & \begin{tabular}[c]{@{}l@{}}$\cdot$ I do not know the definition of the substitute or target word  \\ $\cdot$ I do not know the meaning  of the context sentence\end{tabular}                                                                                                                           \\ \bottomrule
\end{tabular}
\end{adjustbox}
\caption{General annotation guidelines to assign labels to a given triplet $(w, s, w')$. Note that the sentence $s$ can be hard to understand since it is written by beginner L2 English learners.}
\label{tab:ann_rules}
\end{table*}

\begin{table*}[]
\centering
\begin{adjustbox}{max width=\textwidth}
\begin{tabular}{lccl}
\hline
\multicolumn{1}{c}{\textbf{Sentence}}                                                                                                                                                                                                           & \textbf{Substitute} & \textbf{Label} & \multicolumn{1}{c}{\textbf{Reason}}                                                                                                                   \\ \hline
\begin{tabular}[c]{@{}l@{}}For example, I **trust** a person who has \\ more knowledge than I do\end{tabular}                                                                                                                                   & rely on             & 1              & “I trust a person” entails “I rely on a person”                                                                                                       \\ \hline
\begin{tabular}[c]{@{}l@{}}For example, in **private**, an airplane \\ costs 3000 dollars in America from Japan.\end{tabular}                                                                                                                   & personally          & 0              & Unknown meaning of the sentence                                                                                                                       \\ \hline
\begin{tabular}[c]{@{}l@{}}In academic institutions they are introducing new \\ technologies and topics which are going to be used \\ full to the students that are going to change rapidly. \\ Take as an example a **computer**.\end{tabular} & laptop              & 0              & Substitute is a hyponym of target word                                                                                                                \\ \hline
\begin{tabular}[c]{@{}l@{}}If one wants to live a **better** and more plentiful \\ life, I find it basic to experiment, dare a little, risk it \\ a bit and try new things.\end{tabular}                                                        & superior            & 1              & Good substitute - fulfills all criteria                                                                                                               \\ \hline
\begin{tabular}[c]{@{}l@{}}In **addition**, I will argue over the belief in the \\ following reasons.\end{tabular}                                                                                                                              & furthermore         & -1             & Ungrammatical substitute                                                                                                                              \\ \hline
The cow **jumped** over the moon.                                                                                                                                                                                                               & leap                & -1             & \begin{tabular}[c]{@{}l@{}}Ungrammatical (“leap” vs “lept”), verb tense\\ does not match.\end{tabular}                                                \\ \hline
He **grew** up in the town.                                                                                                                                                                                                                     & evolved             & -1             & Highlighed word is part of a phrase (“grow up”)                                                                                                       \\ \hline
\begin{tabular}[c]{@{}l@{}}It's just impossible for him to additionally work \\ **voluntarily**.\end{tabular}                                                                                                                                   & willingly           & 0              & \begin{tabular}[c]{@{}l@{}}Additional context is needed to judge if “voluntarily” \\ refers to “working without pay” or “working freely”\end{tabular} \\ \hline
\begin{tabular}[c]{@{}l@{}}Once the undergraduate studies are **pursued** \\ by a student, the student is more aware of different \\ subjects and the knowledge he has gained in the period \\ of his studies.\end{tabular}                     & embarked on         & 1              & \begin{tabular}[c]{@{}l@{}}Figurative use case of “embarked on”, \\ which conveys the same meaning as “pursued”.\end{tabular}                         \\ \hline
\end{tabular}
\end{adjustbox}
\caption{Example annotations and the reasons behind. Note that for each sentence, the target word is encompassed with double asterisks (**).}
\label{tab:ann_examples}
\end{table*}

\begin{table*}[]
\centering
\begin{adjustbox}{max width=\textwidth}
\begin{tabular}{l}
\hline
\begin{tabular}[c]{@{}l@{}}You are a helpful assistant to perform a lexical substitution task. Specifically, you will be given a tuple of text consisting of \\ 1) context with target word indicated using asterisks, and a 2) natural language query. \\ You should generate exactly five substitutes separated by commas. Do not generate the same word as the target word.\\ Here are some examples:\\ \\ I have completed the invoices for April, May and June and we owe Pasadena each month for a **total** of \$3,615,910.62. \\ I am waiting to hear back from Patti on May and June to make sure they are okay with her.\\ Q: What are appropriate substitutes for **total** in the above text?\\ A: amount, sum, price, balance, gross\\ \\ …I thought as much. Now leave, before I **call** the rats on you.” We left.\\ Q: What are appropriate substitutes for **call** in the above text?\\ A: summon, order, rally, send, sic\\ \\ The e-commerce free **zone** is situated in north Dubai, near the industrial free **zone** in Hebel Ali.\\ Q: What are appropriate substitutes for **zone** in the above text? \\ A: sector, district, area, region, section\\ \\ The state's action, the first in the nation, has the blessing of the American Psychological Association (APA), which considers \\ prescriptive authority a **logical** extension of psychologists' role as health-care providers.\\ Q: What are appropriate substitutes for **logical** in the above text? \\A: rational, reasonable, sensible, justifiable, relevant\\ \\ They nodded. “Oh, **totally**,” said the hunchback. “I get that all the time around here.”\\ Q: What are appropriate substitutes for **totally** in the above text?\\ A: absolutely, for sure, surely, completely, definitely\\ \\ \{Insert CONTEXT with **TARGET** here\}\\ Q: What are appropriate substitutes for **\{TARGET\}** in the above text?\\ A:\end{tabular} \\ \hline
\end{tabular}
\end{adjustbox}
\caption{The prompt for GPT-4 to generate candidate substitutes for human annotation. Note that we incorporate five in-context examples from Swords \cite{lee2021swords}.}
\label{tab:gpt4_prompt_gen}
\end{table*}

\begin{table*}[]
\centering
\begin{adjustbox}{max width=\textwidth}
\begin{tabular}{l}
\hline
\begin{tabular}[c]{@{}l@{}}You are about to perform a lexical substitution task, considering the proficiency level of the substitute compared to the target word in a sentence. \\ The task is to generate a set of candidate substitutes separated by commas for a target word in a given sentence. The target word is highlighted in \\ the sentence, encompassed by two double asterisks.\\ \\ Target word: {[}TARGET{]}\\ Sentence: {[}SENTENE{]}\\ Substitutes:\end{tabular} \\ \hline
\end{tabular}
\end{adjustbox}
\caption{Prompt for zero-shot evaluation of GPT-4 and ChatGPT on ProLex. The prompt takes a given target word and a context sentence, and outputs a list of substitutes separated by commas.}
\label{tab:zero_prompt}
\end{table*}

\begin{table*}[]
\centering
\begin{adjustbox}{max width=\textwidth}
\begin{tabular}{l}
\hline
\begin{tabular}[c]{@{}l@{}}You are about to synthesize data for a lexical substitution task, considering the proficiency level of the substitute compared to the target word in a sentence. \\ Concretely, for each data point, I will first give you an original sentence. Then, you should follow the following steps to create a complete data point: \\ 1) Based on the queried Part of Speech tag, select as unique as possible a content word as the target word to be substituted from the sentence (Content words \\ include nouns, verbs, adjectives and adverbs). Do only select single word as the target; \\ 2) generate at least 10 candidate substitutes (separated by commas) for the selected content word from Step 2;\\ 3) final candidate substitutes after excluding candidates from Step 3 that are not common expressions in actual English use.\\ \\ You should make sure each of the generated substitutes follows exactly the following characteristics: \\ a) does NOT change the meaning and semantics of the sentence, and \\ b) is a common collocation or expression in actual English use (appears at least five times in Corpus of Contemporary American English), and \\ c) in general, matches at least one connotation of the target word in the context sentence, and \\ d) is grammatically correct, and \\ e) has an equal or higher language proficiency level compared to the target word. Please use CEFR (Common European Framework of Reference for Languages) \\ standard to describe the language proficiency of a word. The specification of CEFR levels (from the least proficient to the most proficient) is defined as follows: \\ A1 (beginner), A2 (Elementary), B1 (Intermediate), B2 (Upper Intermediate), C1 (Advanced), C2 (Proficient).\\ \\ Here are some examples (Tagged sentence denotes sentence where the target word is surrounded by two double asterisks). Do not change the original sentence:\\ Query Part of Speech tag: adverb\\ Original Sentence: Students can learn to study independently from understanding ideas and concepts.\\ Tagged Sentence: Students can learn to study **independently** from understanding ideas and concepts.\\ Target word: independently (B2 - Upper Intermediate)\\ Candidate Substitutes: autonomously (C2 - Proficient), individually (C1 - Advanced), solo (B2 - Upper Intermediate)\\ Final Substitutes: autonomously, individually, solo\\ \\ Query Part of Speech tag: adjective\\ Original Sentence:  It is because of this that various kinds of people with special knowledge can complement each other.\\ Tagged Sentence: It is because of this that various kinds of people with **special** knowledge can complement each other.\\ Target word: special (AI - beginner)\\ Candidate Substitutes: specific (A2 - Elementary), distinctive (C1 - Advanced), exclusive (B2 - Upper Intermediate), unique (B2 - Upper Intermediate), particular (A2 - Elementary)\\ Final Substitutes: specific, distinctive, unique, particular\\ \\ Query Part of Speech tag: noun\\ Original Sentence: At the start of a life a person doesn't have success yet, only during life does your action make your success. \\ Tagged Sentence: At the start of a life a person doesn't have success yet, only during life does your **action** make your success. \\ Target word: action (A1 - beginner)\\ Candidate Substitutes: behavior (A2 - Elementary), conduct (B2 - Upper Intermediate), operation (B1 - Intermediate), undertaking (C1 - Advanced), activity (A1 - beginner)\\ Final Substitutes: behavior, conduct, activity\\ \\ Query Part of Speech tag: verb\\ Original Sentence: It has no arguments to support it and is terribly broad.\\ Tagged Sentence: It has no arguments to **support** it and is terribly broad.\\ Target word: support (A2 - Elementary)\\ Candidate Substitutes: back (B2 - Upper Intermediate), substantiate (C2 - Proficient), uphold (C1 - Advanced), justify (B2 - Upper Intermediate)\\ Final Substitutes: back, substantiate, justify\\ \\ Query Part of Speech tag: adverb\\ Original Sentence: many old people have diseases that rob them of their health and make them unable.\\ -\textgreater POS does not exist in the sentence.\\ \\ Please note that if there are no words that correspond to the queried part of speech tag in the original sentence, simply generate "POS does not exist in the sentence".  \\ Do only select single word as the target.\\ \\ Now, please generate:\\ \\ Query Part of Speech tag: {[}Query\_POS{]} \\ Original sentence: {[}Sentence{]}-\textgreater{}\end{tabular} \\ \hline
\end{tabular}
\end{adjustbox}
\caption{Prompt used to synthesize training data for fine-tuning: it generates 1) the target word and tags the target word in the sentence; 2) an initial set of candidate substitutes along with their CEFR proficiency level; and 3) the final proficiency-oriented substitutes.}
\label{tab:gpt4_syn}
\end{table*}

\begin{table*}[]
\centering
\begin{adjustbox}{max width=\textwidth}
\begin{tabular}{l}
\begin{tabular}[c]{@{}l@{}}You are about to perform a lexical substitution task, considering the proficiency level of the substitute compared to the target word in a sentence. \\ The task is to generate a set of candidate substitutes separated by commas for a target word in a given sentence. The target word is highlighted in the sentence, encompassed by two double asterisks. \\ Here are some examples: \\ Target word: honest \\ Sentence: This is not right as people think that the message sent to them is **honest**, thus they believe whatever they hear. \\ Substitutes: truthful, sincere, genuine \\  \\ Target word: fully \\ Sentence: that provides them with access to enjoy their life **fully**. \\ Substitutes: completely, totally, absolutely, wholly \\  \\ Target word: roads \\ Sentence: the government may build more **roads** so that they provide translation. \\ Substitutes: highways, tracks, routes, lanes \\  \\ Target word: backs \\ Sentence: Few year **backs**, music industry was dominated by the Walkman from Sony. \\ Substitutes: ago, previously, earlier, formerly\\ \\ Target word: For example \\ Sentence: **For example**, let' s talk about the French Revolution. \\ Substitutes: For instance, To illustrate \\ \\ Target word: better \\ Sentence: But if you look into it, why do pairs of shoes coast so much, because it is an advertisement that Nike makes their shoes look **better** than everyone else's. \\ Substitutes: superior, improved, exceptional \\  \\ Target word: any \\ Sentence: Learning facts in a scientific field, like for example the resulting speed of an object, does not give **any** clue about the result of an experiment in another context. \\ Substitutes: whatsoever, at all, absolutely \\  \\ Target word: rather \\ Sentence: When they achieve one particular goal they proceed with a new goal **rather** than doing what they already know. \\ Substitutes: preferably, instead, alternatively, ideally \\  \\ Target word: Balancing \\ Sentence: **Balancing** between these methods is actually what we need. \\ Substitutes: Harmonizing \\  \\ Target word: come \\ Sentence: we leave like one day will **come** \\ Substitutes: arrive, appear, emerge \\  \\ ... (22 more examples)\\ \\ Target word: {[}TARGET{]}\\ Sentence: {[}SENTENCE{]}\\ Substitutes:\end{tabular}
\end{tabular}
\end{adjustbox}
\caption{Prompt for in-context learning with GPT-4 and ChatGPT.}
\label{tab:icl_prompt}
\end{table*}

\end{document}